\documentclass[11pt,a4paper]{article}
\usepackage{times,latexsym}
\usepackage{url}
\usepackage[T1]{fontenc}
\usepackage{booktabs}
\usepackage{multirow}
\usepackage{algorithmic}
\usepackage{algorithm}
\usepackage{amsmath,amsfonts,amssymb,mathtools}
\usepackage{tikz-dependency}
\usepackage{graphicx}
\usepackage{color}
\usepackage{colortbl}
\usepackage{bbding}

\newcommand{\mathboldface}[1]{\mbox{\boldmath $#1$}}
\newcommand{\VEC}[1]{\mathboldface{#1}}
\DeclareMathOperator*{\argmax}{arg\,max}

\DeclarePairedDelimiter\norm{\lVert}{\rVert}
\DeclarePairedDelimiter\tuple{\langle}{\rangle}

\usepackage[acceptedWithA]{tacl2018v2}
\usepackage{xspace,mfirstuc,tabulary}

\newif\iftaclinstructions
\taclinstructionsfalse 
\iftaclinstructions
\renewcommand{\confidential}{}
\renewcommand{\anonsubtext}{(No author info supplied here, for consistency with
TACL-submission anonymization requirements)}
\newcommand{\instr}
\fi

\iftaclpubformat 

\else

\fi

\title{Instance-Based Neural Dependency Parsing}

\author{Hiroki Ouchi$^{1,3}$ \hspace{0.7cm} Jun Suzuki$^{2,3}$  \hspace{0.7cm} Sosuke Kobayashi$^{2,4}$ \\
\textbf{Sho Yokoi}$^{2,3}$ \hspace{0.7cm} \textbf{Tatsuki Kuribayashi}$^{2,5}$  \hspace{0.7cm} \textbf{Masashi Yoshikawa}$^{2,3}$  \hspace{0.7cm}\textbf{Kentaro Inui}$^{2,3}$\\
  {$^1$ NAIST} \hspace{0.3cm} {$^2$ Tohoku University} \hspace{0.3cm} {$^3$ RIKEN} \hspace{0.3cm} {$^4$ Preferred Networks, Inc.} \hspace{0.3cm} {$^5$ Langsmith, Inc.}\\
  {\tt hiroki.ouchi@is.naist.jp, sosk@preferred.jp}, \\ \{{\tt jun.suzuki,yokoi,kuribayashi,yoshikawa,inui\}@tohoku.ac.jp}}

\date{}

\begin{document}
\maketitle

\begin{abstract}
Interpretable rationales for model predictions are crucial in practical applications.
We develop neural models that possess an interpretable inference process for dependency parsing.
Our models adopt \textit{instance-based inference}, where dependency edges are extracted and labeled by comparing them to edges in a training set.
The training edges are explicitly used for the predictions; thus, it is easy to grasp the contribution of each edge to the predictions.
Our experiments show that our instance-based models achieve competitive accuracy with standard neural models and have the reasonable plausibility of instance-based explanations.
\end{abstract}

\section{Introduction}
\label{sec:intro}

While deep neural networks have improved prediction accuracy in various tasks, rationales underlying the predictions have been more challenging for humans to understand~\cite{lei-etal-2016-rationalizing}.
In practical applications, interpretable rationales play a crucial role in driving humans' decisions and promoting human-machine cooperation~\cite{ribeiro2016should}.
From this perspective, the utility of \textbf{instance-based learning}~\cite{aha1991instance}, a traditional machine learning method, has been realized again~\cite{papernot2018deep}.

Instance-based learning is a method that learns similarities between training instances and infers a value or class for a test instance on the basis of similarities against the training instances.
On the one hand, standard neural models encode all the knowledge in the parameters, making it challenging to determine what knowledge is stored and used for predictions~\cite{guu2020realm}.
On the other hand, models with instance-based inference explicitly use training instances for predictions and can exhibit the instances that significantly contribute to the predictions.
The instances play a role of an explanation to the question: \textit{why did the model make such a prediction?}
This type of explanation is called \textbf{instance-based explanation}~\cite{caruana1999case,baehrens2010explain,plumb2018model}, which facilitates the users' understandings of model predictions and allows users to make decisions with higher confidence \cite{kolodneer1991improving,ribeiro2016should}.

It is not trivial to combine neural networks with instance-based inference processes while keeping high prediction accuracy.
Recent studies in image recognition seek to develop such methods~\cite{wang2014learning,hoffer2015deep,liu2017sphereface,wang2018cosface,deng2019arcface}.
This paradigm is called deep metric learning.
Compared to image recognition, there are much fewer studies on deep metric learning in natural language processing (NLP).
As a few exceptions, \citet{wiseman-stratos-2019-label} and \citet{ouchi-etal-2020-instance} developed neural models that have an instance-based inference process for sequence labeling tasks.
They reported that their models have high explainability without sacrificing the prediction accuracy.

As a next step from targeting consecutive \textbf{tokens}, we study instance-based neural models for \textbf{relations} between discontinuous elements.
To correctly recognize relations, systems need to capture associations between elements.
As an example of relation recognition, we address dependency parsing, where systems seek to recognize binary relations between tokens (hereafter \textbf{edges}).
Traditionally, dependency parsers have been a useful tool for text analysis.
An unstructured text of interest is parsed, and its structure leads users to a deeper understanding of the text.
By successfully introducing instance-based models to dependency parsing, users can extract dependency edges along with similar edges as a rationale for the parse, which further helps the process of text analysis.

In this paper, we develop new instance-based neural models for dependency parsing, equipped with two inference modes: (i) explainable mode and (ii) fast mode.
In the explainable mode, our models make use of similarities between the candidate edge and each edge in a training set.
By looking at the similarities, users can quickly check which training edges significantly contribute to the prediction.
In the fast mode, our models run as fast as standard neural models, while general instance-based models are much slower than standard neural models because of the dependence on the number of training instances.
The fast mode is motivated by the actual situation: in many cases, users want only predictions, and when the predictions seem suspicious, they want to check the rationales.
So, the fast mode does not offer rationales, but instead, it enables faster parsing that outputs exactly the same predictions as the explainable mode.
Users can freely switch between the explainable and fast modes according to their purposes.
This property is realized by taking advantage of the linearity of score computation in our models and avoids comparing a candidate edge to each training edge one by one for computing the score at test time (see Section~\ref{sec:comp} for details).

Our experiments on multilingual datasets show that our models can achieve competitive accuracy with standard neural models.
In addition, we shed light on the plausibility of instance-based explanations, which has been underinvestigated in dependency parsing.
We verify whether our models meet a minimal requirement related to the plausibility~\cite{hanawa2021evaluation}.
Additional analysis reveals the existence of \textbf{hubs}~\cite{radovanovic2010hubs}, a small number of specific training instances that often appear as nearest neighbors, and that hubs have a terrible effect on the plausibility.
Our main contributions are as follows:
\vspace{-0.2cm}
\begin{itemize}
\setlength{\parskip}{0cm} 
\setlength{\itemsep}{0.1cm}
\item This is the first work to develop and study instance-based neural models\footnote{Our code is publicly available at \url{https://github.com/hiroki13/instance-based-dependency-parsing}} for dependency parsing (Section~\ref{sec:method});
\item Our empirical results show that our instance-based models achieve competitive accuracy with standard neural models (Section~\ref{sec:main-results});
\item Our analysis reveals that L2-normalization for edge representations suppresses the hubs' occurrence, and as a result, succeeds in improving the plausibility of instance-based explanations (Sections~\ref{sec:label-sim}~and~\ref{sec:analysis}).
\end{itemize}

\section{Related Work}
\label{se:related}

\subsection{Dependency Parsing}
There are two major paradigms for dependency parsing~\cite{kubler2009dependency}: (i) \textbf{transition-based} paradigm~\cite{nivre2003efficient,yamada2003statistical} and (ii) \textbf{graph-based} paradigm~\cite{mcdonald-etal-2005-non}.
Recent literature often adopts the graph-based paradigm and achieves high accuracy~\cite{dozat2017deep,zhang-etal-2017-dependency-parsing,hashimoto-etal-2017-joint,clark-etal-2018-semi,ji-etal-2019-graph,zhang-etal-2020-efficient}.
The first-order \textbf{edge-factored} models under this paradigm factorize the score of a dependency tree into independent scores of single edges~\cite{mcdonald-etal-2005-non}.
The score of each edge is computed on the basis of its edge feature.
This decomposable property is preferable for our work because we want to model similarities between single edges.
Thus, we adopt the basic framework of the first-order edge-factored models for our instance-based models.

\subsection{Instance-Based Methods in NLP}
Traditionally, instance-based methods (\textit{memory-based learning}) have been applied to a variety of NLP tasks~\cite{daelemans2005memory}, such as part of speech tagging~\cite{daelemans-etal-1996-mbt}, NER~\cite{tjong-kim-sang-2002-memory,de-meulder-daelemans-2003-memory,hendrickx-van-den-bosch-2003-memory}, partial parsing~\cite{daelemans1999memory,sang2002memory}, phrase-structure parsing~\cite{lebowitz1983memory,scha1999memory,kubler2004memory,bod2009exemplar}, word sense disambiguation~\cite{veenstra2000memory}, semantic role labeling~\cite{akbik-li-2016-k}, and machine translation (MT)~\cite{nagao1984framework,sumita-iida-1991-experiments}.

\citet{nivre-etal-2004-memory} proposed an instance-based (memory-based) method for transition-based dependency parsing.
The subsequent actions of a transition-based parser are selected at each step by comparing the current parser configuration to each of the configurations in the training set.
Here, each parser configuration is treated as an instance and plays a role of rationales for predicted actions but not for predicted edges.
Generally, parser configurations are not directly mapped to each predicted edge one by one, so it is troublesome to interpret which configurations significantly contribute to edge predictions.
By contrast, since we adopt the graph-based one, our models can naturally treat each edge as an instance and exhibit similar edges as rationales for edge predictions.

\subsection{Instance-Based Neural Methods in NLP}
Most of the studies above were published before the current deep learning era.
Very recently, instance-based methods have been revisited and combined with neural models in language modeling~\cite{khandelwal2019generalization}, MT~\cite{khandelwal2020nearest}, and question answering~\cite{lewis2020retrieval}.
They augment a main neural model with a non-parametric sub-module that retrieves auxiliary objects, such as similar tokens and documents.
\citet{guu2020realm} proposed to parameterize and learn the sub-module for a target task.

These studies assume a different setting from ours.
There is no ground-truth supervision signal for retrieval in their setting, so they adopt non-parametric approaches or indirectly train the sub-module to help a main neural model from the supervision signal of the target task.
In our setting, the main neural model plays a role in retrieval and is directly trained with ground-truth objects (annotated dependency edges).
Thus, our findings and insights are orthogonal to theirs.

\subsection{Deep Metric Learning}
Our work can be categorized into the \textit{deep metric learning} research in terms of the methodological perspective.
Although the origins of metric learning can be traced back to some earlier work \cite{short1981optimal,friedman1994flexible,hastie1996discriminant}, the pioneering work is \citet{xing2002distance}.\footnote{If you would like to know the history of metric learning in more detail, please read \citet{bellet2013survey}.}
Since then, many methods using neural networks for metric learning have been proposed and studied.

Deep metric learning methods can be categorized into two classes from the training loss perspective~\cite{sun2020circle}: (i) learning with \textbf{class-level labels} and (ii) learning with \textbf{pair-wise labels}.
Given class-level  labels, the first one learns to classify each training instance to its target class with a classification loss, e.g., Neighbourhood Component Analysis (NCA)~\cite{goldberger2005neighbourhood}, L2-constrained softmax loss~\cite{ranjan2017l2}, SpereFace~\cite{liu2017sphereface}, CosFace~\cite{wang2018cosface}, and ArcFace~\cite{deng2019arcface}.
Given pair-wise labels, the second one learns pair-wise similarity (the similarity between a pair of instances), e.g., contrastive loss~\cite{hadsell2006dimensionality}, triplet loss~\cite{wang2014learning,hoffer2015deep}, N-pair loss~\cite{sohn2016improved}, and Multi-similarity loss~\cite{wang2019multi}.
Our method is categorized into the first one because it adopts a classification loss (Section~\ref{sec:method}).

\subsection{Neural Models Closely Related to Ours}
Among the metric learning methods above, NCA~\cite{goldberger2005neighbourhood} shares the same spirit as our models.
In this framework, models learn to map instances with the same label to the neighborhood in a feature space.
\citet{wiseman-stratos-2019-label} and \citet{ouchi-etal-2020-instance} developed NCA-based neural models for sequence labeling.
We discuss the differences between their models and ours later in more detail (Section~\ref{sec:method:discussion}).

\section{Dependency Parsing Framework}
\label{sec:problem}
We adopt a two-stage approach \cite{mcdonald-etal-2006-multilingual,zhang-etal-2017-dependency-parsing}: we first identify dependency edges (unlabeled dependency parsing) and then classify the identified edges (labeled dependency parsing).
More specifically, we solve edge identification as head selection and solve edge classification as multi-class classification.\footnote{Although some previous studies adopt multi-task learning methods for edge identification and classification tasks \cite{dozat2017deep,hashimoto-etal-2017-joint}, we independently train a model for each task because the interaction effects produced by multi-task learning make it challenging to analyze models' behaviors.}

\subsection{Edge Identification}
To identify unlabeled edges, we adopt the \textbf{head selection} approach~\cite{zhang-etal-2017-dependency-parsing}, in which a model learns to select the correct head of each token in a sentence.
This simple approach enables us to train accurate parsing models in a GPU-friendly way.
We learn the representation for each edge to be discriminative for identifying correct heads.

Let $\VEC x = (x_0, x_1, \dots, x^{}_T)$ denote a tokenized input sentence, where $x_0$ is a special \textsc{root} token and $x_1, \dots, x^{}_T$ are original $T$ tokens, and $\langle x_j, x_i \rangle$ denote an edge from head token $x_j$ to dependent token $x_i$.
We define the probability of token $x_j$ being the head of token $x_i$ in the sentence $\VEC x$ as:
\begin{align}
	P(x_j | x_i) = \frac{\exp(s_{\mathrm{head}}(x_j, x_i))}{\sum^T_{k=0} \exp(s_{\mathrm{head}}(x_k, x_i))}
	\label{eq:head-prob}
	\text{.}
\end{align}

\noindent
Here, the scoring function $s_{\mathrm{head}}$ can be any neural network-based scoring function (see Section~\ref{sec:head:scoring}).

At inference time, we choose the most likely head $\hat{y}_i$ for each token $x_i$\footnote{While this greedy formulation has no guarantee to produce well-formed trees, we can produce well-formed ones by using the Chu-Liu-Edmonds algorithm in the same way as \citet{zhang-etal-2017-dependency-parsing}. In this work, we would like to focus on the representation for each edge and evaluate the goodness of the learned edge representation one by one. With such a motivation, we adopt the greedy formulation.}:
\begin{align}
	\hat{y}_i = \argmax_{\mathclap{x_k \colon 0 \le k \le T}} P(x_k | x_i)
	\text{.}
\end{align}

At training time, we minimize the negative log-likelihood of training data:
\begin{equation}
	L = - \sum^{\lvert \mathcal{D} \rvert}_{n=1} \sum_{i=1}^{T^{(n)}} \log P(y^{(n)}_i | x^{(n)}_i)
	\label{eq:head-loss}
	\text{.}
\end{equation}

\noindent
Here, $\mathcal{D} = \{ \VEC x^{(n)}, \VEC y^{(n)}, \VEC r^{(n)} \}^{\lvert \mathcal{D} \rvert}_{n=1}$ is a training set, where $x^{(n)}_i \in \VEC x^{(n)}$ is each input token, $y^{(n)}_i \in \VEC y^{(n)}$ is the gold (ground-truth) head for token $x^{(n)}_i$, and $r^{(n)}_i \in \VEC r^{(n)}$ is the label for edge $\tuple{y^{(n)}_i, x^{(n)}_i}$.

\subsection{Label Classification}
We adopt a simple multi-class classification approach for labeling each unlabeled edge.
We define the probability that each of all possible labels $r \in \mathcal{R}$ will be assigned to the edge $\tuple{x_j, x_i}$:
\begin{align}
	P(r | x_j, x_i) = \frac{\exp(s_{\mathrm{label}}(r, x_j, x_i))}{\displaystyle \sum_{r' \in \mathcal{R}} \exp(s_{\mathrm{label}}(r', x_j, x_i))} 
	\label{eq:label-prob}
	\text{.}
\end{align}

\noindent
Here, the scoring function $s_{\mathrm{label}}$ can be any neural network-based scoring function (see Section~\ref{sec:label:scoring}).

At inference time, we choose the most likely class label from the set of all possible labels $\mathcal{R}$:
\begin{align}
	\hat{r} = \argmax_{\mathclap{r \in \mathcal{R}}} P(r | \hat{y}_i, x_i)
	\text{.}
\end{align}

\noindent
Here, $\hat{y}_i$ is the head token identified by a head selection model.

At training time, we minimize the negative log-likelihood of training data:
\begin{align}
	L = - \sum^{\lvert \mathcal{D} \rvert}_{n=1} \sum_{i =1}^{T^{(n)}} \log P(r^{(n)}_i | y^{(n)}_i, x^{(n)}_i)
	\label{eq:label-loss}
	\text{.}
\end{align} 

\noindent
Here, $r^{(n)}_i \in \mathcal{R}$ is the gold (ground-truth) relation label for gold edge $\tuple{y^{(n)}_i, x^{(n)}_i}$.

\section{Instance-Based Scoring Methods}
\label{sec:method}

For the scoring functions in Eqs.~\ref{eq:head-prob}~and~\ref{eq:label-prob}, we describe our proposed instance-based models.

\subsection{Edge Scoring}
\label{sec:head:scoring}
We would like to assign a higher score to the correct edge than other candidates (Eq.~\ref{eq:head-prob}).
Here, we compute similarities between each candidate edge and ground-truth edges in a training set (hereafter \textit{training edge}).
By summing the similarities, we then obtain the score that indicates how likely the candidate edge is the correct one.

Specifically, we first construct a set of training edges, called the \textit{support set}, $\mathcal{A}(\mathcal{D})$:
\begin{align}
	\nonumber \mathcal{A}(\mathcal{D}) = \{ & \tuple{y_i, x_i} \mid x_i \in \VEC x, y_i \in \VEC y, \\
	& (\VEC x, \VEC y, \VEC r) \in \mathcal{D}  \}
	\text{.}
	\label{eq:head:repset}
\end{align}

\noindent
Here, $y_i$ is the ground-truth head token of token $x_i$.
We then compute and sum similarities between a candidate edge and each edge in the support set:
\begin{align}
	s_\mathrm{head}(x_j, x_i) & = \sum_{\mathclap{\tuple{x_\ell, x_k} \in \mathcal{A}(\mathcal{D})}} \mathrm{sim}(\VEC h^{}_{\tuple{j, i}}, \VEC h^{}_{\tuple{\ell, k}})
	\label{eq:instance-based}
	\text{.}
\end{align}

\noindent
Here, $\VEC h^{}_{\tuple{j, i}}, \VEC h^{}_{\tuple{\ell, k}} \in \mathbb{R}^d$ are $d$-dimensional edge representations (Section~\ref{sec:edge-rep}), and $\mathrm{sim}$ is a similarity function.
Following recent studies of deep metric learning, we adopt the dot product and the cosine similarity:
\begin{align}
	\nonumber \mathrm{sim}_{\mathrm{dot}}(\VEC a, \VEC b) & = \VEC a^\top \VEC b
	\text{,}\\
	\nonumber \mathrm{sim}_{\cos}(\VEC a , \VEC b) & = \tau {\VEC a}^\top_{\mathrm u} {\VEC b}^{}_{\mathrm u}
	\text{,} \\
	\nonumber {\VEC a}^{}_{\mathrm u} & = \VEC a / \norm{\VEC a}
	\text{,} \\
	\nonumber {\VEC b}^{}_{\mathrm u} & = \VEC b / \norm{\VEC b}
	\text{.}
\end{align}

\noindent
As you can see, the cosine similarity is the same as the dot product between two unit vectors: i.e., $\norm{{\VEC a}^{}_{\mathrm u}} = \norm{{\VEC b}^{}_{\mathrm u}} = 1$.
As we will discuss later in Section~\ref{sec:analysis}, this property suppresses the occurrence of \textit{hubs}, compared with the dot product between unnormalized vectors.
Note that, following the previous studies of deep metric learning~\cite{wang2018cosface,deng2019arcface}, we rescale the cosine similarity by using the scaling factor $\tau$ $(0 \leq \tau)$, which works as the temperature parameter in the softmax function.\footnote{In our preliminary experiments, we set $\tau$ by selecting a value from $\{16, 32, 64, 128\}$. As a result, whichever we chose, the prediction accuracy was stably better than $\tau = 1$.}

\subsection{Label Scoring}
\label{sec:label:scoring}
Similarly to the scoring function above, we also design our instance-based label scoring function $s_\mathrm{label}$ in Eq.~\ref{eq:label-prob}.
We first construct a support set $\mathcal{A}(\mathcal{D}; r)$ for each relation label $r \in \mathcal{R}$:
\begin{align}
	\nonumber \mathcal{A}(\mathcal{D}; r) = \{ & \tuple{y_i, x_i} \mid x_i \in \VEC x, y_i \in \VEC y, r_i \in \VEC r, \\
	& (\VEC x, \VEC y, \VEC r) \in \mathcal{D}, r_i = r  \}
	\text{.}
	\label{eq:label:repset}
\end{align}

\noindent
Here, only the edges with label $r$ are collected from the training set.
We then compute and sum similarities between a candidate edge and each edge of the support set:
\begin{align}
\label{eq:label:instance-based}
	s_\mathrm{label}(r, x_j, x_i) =  \sum_{\mathclap{\tuple{x_\ell, x_k} \in \mathcal{A}(\mathcal{D}; r)}} \mathrm{sim}(\VEC h^{}_{\tuple{j, i}}, \VEC h^{}_{\tuple{\ell, k}})
	\text{.}
\end{align}

\noindent
Here is the intuition: if the edge is more similar to the edges with label $r$ than those with other labels, the edge is more likely to have the label $r$.

\subsection{Edge Representation}
\label{sec:edge-rep}
In the proposed models (Eqs.~\ref{eq:instance-based}~and~\ref{eq:label:instance-based}), we use $d$-dimensional edge representations.
We define the representation for each edge $\tuple{x_j, x_i}$ as follows:
\begin{align}
\label{eq:rep}
	\VEC h^{}_{\langle j, i \rangle} = f(\VEC h^\mathrm{dep}_i, \VEC h^\mathrm{head}_j)
	\text{.}
\end{align}

\noindent
Here, $\VEC h^\mathrm{dep}, \VEC h^\mathrm{head} \in \mathbb{R}^d$ are $d$-dimensional feature vectors for the dependent and head, respectively.
These vectors are created from a neural encoder (Section~\ref{sec:encoder}).
When designing $f$, it is desirable to capture the interaction between the two vectors.
By referring to the insights into feature representations of relations on knowledge bases~\cite{bordes2013translating,yang2014embedding,nickel2016holographic}, we adopt a \textit{multiplicative} composition, a major composition technique for two vectors\footnote{In our preliminary experiments, we also tried an \textit{additive} composition and the concatenation of the two vectors. The accuracies by these techniques for unlabeled dependency parsing, however, were both about $20$\%, which is much inferior to that by the multiplicative composition.}:
\begin{align}
	f^{}(\VEC h^\mathrm{dep}_i, \VEC h^\mathrm{head}_j) := \VEC W (\VEC h^\mathrm{dep}_i \odot \VEC h^\mathrm{head}_j)
	\text{.}
	\nonumber
\end{align}

\noindent
Here, the interaction between $\VEC h^\mathrm{dep}_i$ and $\VEC h^\mathrm{head}_j$ is captured by element-wise multiplication $\odot$.
These are composed into one vector, which is then transformed  by a weight matrix $\VEC W \in \mathbb{R}^{d \times d}$ into $\VEC h^{}_{\langle j, i \rangle}$.

\subsection{Fast Mode}
\label{sec:comp}

Do users want rationales for all the predictions?
Maybe not.
In many cases, all they want to do is to parse sentences as fast as possible.
Only when they find a suspicious prediction, they will check the rationale for it.
To fulfill the demand, our parser provides two modes: (i) \textbf{explainable mode} and (ii) \textbf{fast mode}.
The explainable mode, as described in the previous subsections, enables to exhibit similar training instances as rationales, but its time complexity depends on the size of the training set.
By contrast, the fast mode does not provide rationales, but instead, it enables faster parsing than the explainable mode and outputs exactly the same predictions as the explainable mode.
Thus, at test time, users can freely switch between the modes: e.g., they first use the fast mode, and if they find a suspicious prediction, then they will use the explainable mode to obtain the rationale for it.

Formally, if using the dot product and cosine similarity for similarity function in Eq.~\ref{eq:instance-based}, the explainable mode can be rewritten as the fast mode:
\begin{align}
	s_\mathrm{head}(x_j, x_i) & = \sum_{\mathclap{\tuple{x_\ell, x_k} \in \mathcal{A}(\mathcal{D})}} \VEC h^\top_{\tuple{j, i}} \VEC h_{\tuple{\ell, k}} 
	\nonumber \\
	& = \VEC h^\top_{\tuple{j, i}} \sum_{\mathclap{\tuple{x_\ell, x_k} \in \mathcal{A}(\mathcal{D})}} \VEC h^{}_{\tuple{\ell, k}}
	\nonumber \\
	& = \VEC h^\top_{\tuple{j, i}} \VEC h^\mathrm{sum}_{\mathcal{A}(\mathcal{D})}
	\text{,}
	\label{eq:fast}
\end{align}

\noindent
where $\VEC h^\mathrm{sum}_{\mathcal{A}(\mathcal{D})} := \sum_{\tuple{x_\ell, x_k} \in \mathcal{A}(\mathcal{D})} \VEC h^{}_{\tuple{\ell, k}}$.
In this way, once you sum all the vectors in the training set $\VEC h^{}_{\tuple{\ell, k}} \: (\tuple{x_\ell, x_k} \in \mathcal{A}(\mathcal{D}))$, you can reuse the summed vector without searching the training set again.
At test time, you can precompute this summed vector $\VEC h^\mathrm{sum}_{\mathcal{A}(\mathcal{D})}$ before running the model on a test set, which reduces the exhaustive similarity computation over the training set to the simple dot product between the two vectors $\VEC h^\top_{\tuple{j, i}} \VEC h^\mathrm{sum}_{\mathcal{A}(\mathcal{D})}$.\footnote{In the same way as Eq.~\ref{eq:fast}, we can transform  $s_\mathrm{label}$ in Eq.~\ref{eq:label:instance-based} to the fast mode.}

\subsection{Relations to Existing Models}
\label{sec:method:discussion}

\paragraph{The closest models to ours.}
\citet{wiseman-stratos-2019-label} and \citet{ouchi-etal-2020-instance} proposed an instance-based model using Neighbourhood Components Analysis (NCA)~\cite{goldberger2005neighbourhood} for sequence labeling.
Given an input sentence of $T$ tokens, $\VEC x = (x_1, \dots, x^{}_T)$, the model first computes the probability that a token (or span) $x_i \in \VEC x$ in the sentence selects each of all the tokens in the training set $x_j \in \VEC x^{}_\mathcal{D}$ as its neighbor: 
\begin{align}
	P(x_j | x_i) & = \frac{\exp(\mathrm{sim}(x_j, x_i))}{{\displaystyle \sum_{x_k \in \VEC x^{}_\mathcal{D}}} \exp(\mathrm{sim}(x_k, x_i))}
	\text{.}
	\nonumber
\end{align}

\noindent
The model then constructs a set of only the tokens $x_j$ associated with a label $y$: $\mathcal{X}(\VEC x^{}_\mathcal{D}; y) = \{x_j \mid  x_j \in \VEC x^{}_\mathcal{D}, y_j = y \}$, and computes the probability that each token $x_i$ will be assigned a label $y$:
\begin{align}
	P(y | x_i) = {\displaystyle \sum_{x_j \in \mathcal{X}(\mathcal{D}; y)}} P(x_j, x_i)
	\text{.}
	\nonumber
\end{align}

\noindent
The point is that while our models first sums the similarities (Eq~\ref{eq:instance-based}) and then put the summed score into exponential form as $\exp(s_{\mathrm{head}}(x_j, x_i))$, their model puts each similarity into exponential form as $\exp(\mathrm{sim}(x_k, x_i))$ before the summation.
The different order of using exponential function makes it impossible to rewrite their model as the fast mode, so their model always has to compare a token $x_i$ to each of the training set $x_j \in \VEC x^{}_\mathcal{D}$.
This is the biggest difference between their model and ours.
While we leave the performance comparison between the NCA-based models and ours for future work, our models have an advantage over the NCA-based models in that our models offer two options, the explainable and fast modes.

\paragraph{Standard models using weights.}
Typically, neural models use the following scoring functions:
\begin{align}
	s_\mathrm{head}(x_j,x_i) & = \VEC w^\top \VEC h_{\tuple{j, i}}
	\text{,}
	\label{eq:head:weight-based}\\
	s_\mathrm{label}(r, x_j, x_i) & = \VEC w^\top_r \VEC h_{\tuple{j, i}}
	\text{.}
	\label{eq:label:weight-based}
\end{align}

\noindent
Here, $\VEC w \in \mathbb{R}^d$ is a learnable weight vector and $\VEC w_r \in \mathbb{R}^d$ is a learnable weight vector associated with label $r \in \mathcal{R}$.
In previous work~\cite{zhang-etal-2017-dependency-parsing}, this form is used for dependency parsing.
We call such models \textbf{weight-based} models.
\citet{caruana1999case} proposed to combine weight-based models with instance-based inference: at inference time, the weights are discarded, and only the trained encoder is used to extract feature representations for instance-based inference.
Such combination has been reported to be effective for image recognition~\cite{ranjan2017l2,liu2017sphereface,wang2018cosface,deng2019arcface}.
In dependency parsing, there has been no investigation on it.
Since such a combination can be a promising method, we investigate its utility (Section~\ref{sec:results}).

\section{Experimental Setup}
\label{sec:exp}

\subsection{Data}
\begin{table}[t]
  \centering
  {\small
  \begin{tabular}{l | c | c | c | r} \toprule
                 Language & Treebank & Family & Order & Train \\ \midrule
                 Arabic & PADT & non-IE & VSO & 6.1k \\
                 Basque & BDT & non-IE & SOV & 5.4k \\
                 Chinese & GSD & non-IE & SVO & 4.0k \\
                 English & EWT & IE & SVO & 12.5k \\
                 Finnish & TDT & non-IE & SVO & 12.2k \\
                 Hebrew & HTB & non-IE & SVO & 5.2k \\
                 Hindi & HDTB & IE & SOV & 13.3k \\
                 Italian & ISDT & IE & SVO & 13.1k \\
                 Japanese & GSD & non-IE & SOV & 7.1k \\
                 Korean & GSD & non-IE & SOV & 4.4k \\
                 Russian & SynTagRus & IE & SVO & 48.8k \\
                 Swedish & Talbanken & IE & SVO & 4.3k \\
                 Turkish & IMST & non-IE & SOV & 3.7k \\ \bottomrule
  \end{tabular}
  }
  \vspace{-0.2cm}
  \caption{Dataset information. ``Family'' indicates if Indo-European (IE) or not. ``Order'' indicates dominant word orders according to WALS~\cite{haspelmath2005world}. ``Train'' is the number of training sentences.}
  \label{tab:data-stats}
\end{table}

We use English PennTreebank (PTB)~\cite{marcus-etal-1993-building} and Universal Dependencies (UD)~\cite{mcdonald-etal-2013-universal}.
Following previous studies~\cite{kulmizev2019deep,smith2018investigation,de2017old}, we choose a variety of 13 languages\footnote{These languages have been selected by considering the perspectives of different language families, different morphological complexity, different training sizes and domains.} from the UD v2.7.
Table 1 shows information about each dataset.
We follow the standard training-development-test splits.

\subsection{Neural Encoder Architecture}
\label{sec:encoder}
To compute $\VEC h^\mathrm{dep}$ and $\VEC h^\mathrm{head}$ (in Eq.~\ref{eq:rep}), we adopt the encoder architecture proposed by \citet{dozat2017deep}.
First, we map the input sequence $\VEC x = (x_0, \dots, x_T)$\footnote{We use the gold tokenized sequences in PTB and UD.} to a sequence of token representations, $\VEC h^\mathrm{token}_{0:T} = (\VEC h^\mathrm{token}_0, \dots, \VEC h^\mathrm{token}_T)$, each of which is $\VEC h^\mathrm{token}_t = [\VEC e_t; \VEC c_t; \VEC b_t]$, where $\VEC e_t$, $\VEC c_t$, and $\VEC b_t$ are computed by word embeddings\footnote{For PTB, we use 300 dimensional GloVe~\cite{pennington-etal-2014-glove}. For UD, we use 300 dimensional fastText~\cite{grave2018learning}. During training, we fix them.}, character-level CNN, and BERT~\cite{devlin-etal-2019-bert}\footnote{We first conduct subword segmentation for each token of the input sequence. Then, the BERT encoder takes as input the subword-segmented sequences and computes the representation for each subword. Here, we use the (last layer) representation of the first subword within each token as its token representation. For PTB, we use ``BERT-Base, Cased.'' For UD, we use ``BERT-Base, Multilingual Cased.''}, respectively.
Second, the sequence $\VEC h^\mathrm{token}_{0:T}$ is fed to bidirectional LSTM (BiLSTM)~\cite{graves:13} for computing contextual ones: $\VEC h^\mathrm{lstm}_{0:T} = (\VEC h^\mathrm{lstm}_0, \dots, \VEC h^\mathrm{lstm}_T) = \mathrm{BiLSTM}(\VEC h^\mathrm{token}_{0:T} )$.
Finally, $\VEC h^\mathrm{lstm}_t \in \mathbb{R}^{2d}$ is transformed as $\VEC h^\mathrm{dep}_t = \VEC W^\mathrm{dep} \VEC h^\mathrm{lstm}_t$ and $\VEC h^\mathrm{head}_t = \VEC W^\mathrm{head} \VEC h^\mathrm{lstm}_t$, where $\VEC W^\mathrm{dep} \in \mathbb{R}^{d \times 2d}$ and $\VEC W^\mathrm{head} \in \mathbb{R}^{d \times 2d}$ are parameter matrices.

\subsection{Mini-Batching}
We train models with the mini-batch stochastic gradient descent method.
To make the current mini-batch at each time step, we follow a standard technique for training instance-based models~\cite{hadsell2006dimensionality,oord2018representation}.

At training time, we make the mini-batch that consists of query and support sentences at each time step.
A model encodes the sentences and the edge representations used for computing similarities between each candidate edge in the query sentences and each gold edge in the support sentences.
Here, due to the memory limitation of GPUs, we randomly sample a subset from the training set at each time step: i.e.,  $(\VEC x^{(n)}, \VEC y^{(n)}, \VEC r^{(n)}) \sim \mathrm{Uniform}(\mathcal{D})$.
In edge identification, for query sentences, we randomly sample a subset $\mathcal{D}'_\mathrm{q}$ of $N$ sentences from $\mathcal{D}$.
For support sentences, we randomly sample a subset $\mathcal{D}'_\mathrm{s}$ of $M$ sentences from $\mathcal{D}$, and construct and use the support set $\mathcal{A}(\mathcal{D}'_\mathrm{s})$ instead of $\mathcal{A}(\mathcal{D})$ in Eq.~\ref{eq:head:repset}.
In label classification, we would like to guarantee that the support set in every mini-batch always contains at least one edge for each label.
To do so, we randomly sample a subset $\mathcal{A}'(\mathcal{D}; r)$ of $U$ edges from the support set for each label~$r$: i.e., $\tuple{y^{(n)}_i, x^{(n)}_i} \sim \mathrm{Uniform}(\mathcal{A}(\mathcal{D}; r))$ in Eq.~\ref{eq:label:repset}.
Note that each edge $\tuple{y^{(n)}_i, x^{(n)}_i}$ is in the $n$-th sentence $\VEC x^{(n)}$ in the training set $\mathcal{D}$, so we put the sentence $\VEC x^{(n)}$ into the mini-batch to compute the representation for $\tuple{y^{(n)}_i, x^{(n)}_i}$.
Actually, we use $N=32$ query sentences in both edge identification and label classification, $M=10$ support sentences in edge identification\footnote{As a result, the whole mini-batch size is $32 + 10 = 42$.}, and $U=1$ support edge (sentence) for each label in label classification\footnote{When $U=1$, the whole mini-batch size is $32 + \lvert \mathcal{R} \rvert$.}.

At test time, we encode each test (query) sentence and compute the representation for each candidate edge on-the-fly.
The representation is then compared to the precomputed support edge representation, $\VEC h^\mathrm{sum}_{\mathcal{A}(\mathcal{D})}$ in Eq~\ref{eq:fast}.
To precompute $\VEC h^\mathrm{sum}_{\mathcal{A}(\mathcal{D})}$, we first encode all the training sentences and obtain the edge representations.
Then, in edge identification, we sum all of them and obtain one support edge representation $\VEC h^\mathrm{sum}_{\mathcal{A}(\mathcal{D})}$.
In label classification, similarly to $\VEC h^\mathrm{sum}_{\mathcal{A}(\mathcal{D})}$, we sum only the edge representations with label~$r$ and obtain one support representation for each label $\VEC h^\mathrm{sum}_{\mathcal{A}(\mathcal{D}; r)}$\footnote{The total number of the support edge representations is equal to the size of the label set $\lvert \mathcal{R }\rvert$.}.

\subsection{Training Configuration}
\begin{table}[t]
  \begin{center}
  {\small
  \begin{tabular}{lr} \toprule
    Name & Value \\ \midrule
    Word Embedding & GloVe (PTB) / fastText (UD)\\
    BERT& BERT-Base\\
    CNN window size & 3\\
    CNN filters & 30 \\
    BiLSTM layers & 2 \\
    BiLSTM units  & 300 dimensions \\
    Optimization & Adam \\
    Learning rate & 0.001 \\
    Rescaling factor $\tau$ & 64 \\
    Dropout ratio & \{0.1, 0.2, 0.3\} \\ \bottomrule
  \end{tabular}
  }
  \end{center}
  \caption{Hyperparameters used in the experiments.}
\label{tab:hyperparam}
\end{table}

Table~\ref{tab:hyperparam} lists the hyperparameters.
To optimize the parameters, we use Adam~\cite{kingma:14} with $\beta_1 = 0.9$ and $\beta_2 = 0.999$.
The initial learning rate is $\eta_0 = 0.001$ and is updated on each epoch as $\eta_t = \eta_0 / (1+\rho t)$, where $\rho = 0.05$ and $t$ is the epoch number completed.
A gradient clipping value is $5.0$~\cite{pascanu2013difficulty}.
The number of training epochs is $100$.
We save the parameters that achieve the best score on each development set and evaluate them on each test set.
It takes less than one day to train on a single GPU, NVIDIA DGX-1 with Tesla V100.

\section{Results and Discussion}
\label{sec:results}

\subsection{Prediction Accuracy on Benchmark Tests}
\label{sec:main-results}
\begin{table*}[t]
  \centering
  {\small
  \begin{tabular}{l | c | cc | cc | cc} \toprule
  		   Learning & Weight-based & \multicolumn{4}{c|}{Weight-based} & \multicolumn{2}{c}{Instance-based} \\ 
                 Inference & Weight-based & \multicolumn{2}{c|}{Weight-based} & \multicolumn{2}{c|}{Instance-based} & \multicolumn{2}{c}{Instance-based} \\
		   Similarity & $\mathrm{dot}$ & $\mathrm{dot}$ & $\cos$ & $\mathrm{dot}$ & $\cos$ & $\mathrm{dot}$ & $\cos$ \\ \midrule
		   System ID & \footnotesize{Kulmizev+'19} & \texttt{WWd} & \texttt{WWc} & \texttt{WId} & \texttt{WIc} & \texttt{IId} & \texttt{IIc} \\ \midrule
PTB-English & -- & 96.4/95.3 & 96.4/95.3 & 96.4/94.4 & 93.0/91.8 & 96.4/95.3 & 96.4/95.3 \\		   
UD-Average & -- /84.9 & 89.0/85.6 & 89.0/85.6 & 89.0/85.2 & 83.0/79.5 & 89.3/85.7 & 89.0/85.5 \\  \midrule
UD-Arabic & -- /81.8 & 87.8/82.1 & 87.8/82.1 & 87.8/81.6 & 84.9/79.0 & 88.0/82.1 & 87.6/81.9 \\ 
UD-Basque & -- /79.8 & 84.9/81.1 & 84.9/80.9 & 84.9/80.6 & 82.0/77.9 & 85.1/80.9 & 85.0/80.8 \\ 
UD-Chinese & -- /83.4 & 85.6/82.3 & 85.8/82.4 & 85.7/81.6 & 80.9/77.3 & 86.3/82.8 & 85.9/82.5 \\ 
UD-English & -- /87.6 & 90.9/88.1 & 90.7/88.0 & 90.9/87.8 & 88.1/85.3 & 91.1/88.3 & 91.0/88.2 \\ 
UD-Finnish & -- /83.9 & 89.4/86.6 & 89.1/86.3 & 89.3/86.1 & 84.1/81.2 & 89.6/86.6 & 89.4/86.4 \\ 
UD-Hebrew & -- /85.9 & 89.4/86.4 & 89.5/86.5 & 89.4/85.9 & 82.7/79.7 & 89.8/86.7 & 89.6/86.6 \\ 
UD-Hindi & -- /90.8 & 94.8/91.7 & 94.8/91.7 & 94.8/91.4 & 91.4/88.0 & 94.9/91.8 & 94.9/91.6 \\ 
UD-Italian & -- /91.7 & 94.1/92.0 & 94.2/92.1 & 94.1/91.9 & 91.5/89.4 & 94.3/92.2 & 94.1/92.0 \\ 
UD-Japanese & -- /92.1 & 94.3/92.8 & 94.5/93.0 & 94.3/92.7 & 92.5/90.9 & 94.6/93.1 & 94.4/92.8 \\ 
UD-Korean & -- /84.2 & 88.0/84.4 & 87.9/84.3 & 88.0/84.2 & 84.3/80.4 & 88.1/84.4 & 88.2/84.5 \\ 
UD-Russian & -- /91.0 & 94.2/92.7 & 94.1/92.7 & 94.2/92.4 & 57.7/56.5 & 94.3/92.8 & 94.1/92.6 \\ 
UD-Swedish &  -- /86.9 & 90.3/87.6 & 90.3/87.5 & 90.4/87.1 & 88.6/85.8 & 90.5/87.5 & 90.4/87.5 \\ 
UD-Turkish & -- /64.9 & 73.0/65.3 & 73.2/65.4 & 73.1/64.5 & 69.9/61.9 & 73.7/65.5 & 72.9/64.7 \\ 
\bottomrule 
  \end{tabular}
  }
  \caption{Comparison between weight-based and instance-based systems. Cells show unlabeled attachment scores (UAS) before the slash and labeled attachment scores (LAS) after the slash on each test set. System IDs stand for the first letters of the options: e.g., \texttt{WId} stands for ``W''eight-based learning and ``I''nstance-based inference using the ``d''ot product. The system ID, Kulmizev+'19, is the graph-based parser with BERT in \citet{kulmizev2019deep}.}
  \label{tab:main-result}
\end{table*}

\begin{table}[t]
  \centering
  {\small
  \begin{tabular}{r | rr | rr} \toprule
                  & \multicolumn{2}{c|}{Weight-Based} & \multicolumn{2}{c}{Instance-Based} \\
                 & $\mathrm{dot}$ & $\cos$ & $\mathrm{dot}$ & $\cos$ \\
                 & \texttt{WWd} & \texttt{WWc} & \texttt{IId} & \texttt{IIc} \\ \midrule
                  \texttt{Emails} &   81.7 & 81.7 & 81.6 &  81.4 \\
                  \texttt{Newsgroups} & 83.1 & 83.3 & 83.1 &  82.9 \\
                  \texttt{Reviews}  & 88.5 & 88.7 & 88.7 &  88.8 \\
                  \texttt{Weblogs} & 81.9 & 80.9 & 80.9 & 81.9 \\ \midrule
                  Average & 83.8 & 83.7 & 83.6 & 83.8 \\
                 \bottomrule
  \end{tabular}
  }
  \caption{UAS in out-of-domain settings, where each model is trained on the source domain ``Yahoo!~Answers" and tested on each of the four target domains.}
  \label{tab:domain-adaptation}
\end{table}

\begin{table}[t]
  \centering
  {\small
  \begin{tabular}{r | rrrr} \toprule
                  & \multicolumn{4}{c}{$M$} \\
                 & $1$ &$10$ & $100$ & ALL \\ \midrule
                  \texttt{Emails} &   81.5 & 81.4 & 81.5 &  81.5 \\
                  \texttt{Newsgroups} & 82.8 & 83.0 & 82.9 &  82.9 \\
                  \texttt{Reviews}  & 88.7 & 88.7 & 88.8 &  88.8 \\
                  \texttt{Weblogs} & 81.8 & 82.1 & 82.0 & 81.9 \\ \midrule
                  Average & 83.7 & 83.8 & 83.8 & 83.8 \\
                 \bottomrule
  \end{tabular}
  }
  \caption{UAS by the instance-based system using the cosine similarity (\texttt{IIc}) and randomly sampled $M$ support training sentences.}
  \label{tab:domain-adaptation-diff-M}
\end{table}

We report averaged unlabeled attachment scores (UAS) and labeled attachment scores (LAS) across three different runs of the model training with random seeds.
We compare 6 systems, each of which consists of two models for edge identification and label classification, respectively.
For reference, we list the results by the graph-based parser with BERT in \citet{kulmizev2019deep}, whose architecture is the most similar to ours.

Table~\ref{tab:main-result} shows UAS and LAS by these systems.
The systems \texttt{WWd} and \texttt{WWc} are the standard ones that consistently use the weight-based scores (Eqs.~\ref{eq:head:weight-based}~and~\ref{eq:label:weight-based}) during learning and inference.
Between these systems, the difference of the similarity functions does not make a gap in the accuracies.
In other words, the dot product and the cosine similarity are on par in terms of the accuracies.
The systems \texttt{WId} and \texttt{WIc} use the weight-based scores during learning and the instance-based ones during inference.
While the system \texttt{WId} using $\mathrm{dot}$ achieved competitive UAS and LAS to those by the standard weight-based system \texttt{WWd}, the system \texttt{WIc} using $\cos$ achieved lower accuracies than those by the system \texttt{WWc}.
The systems \texttt{IId} and \texttt{IIc} consistently use the instance-based scores during learning and inference.
Both of them succeeded in keeping competitive accuracies with those by the standard weight-based ones \texttt{WWd} and \texttt{WWc}.

\paragraph{Out-of-domain robustness.}
We evaluate the robustness of our instance-based models in out-of-domain settings by using the five domains of UD-English: we train each model on the training set of the source domain ``Yahoo!~Answers'' and test it on each test set of the target domains, \texttt{Emails}, \texttt{Newsgroups}, \texttt{Reviews} and \texttt{Weblogs}.
As Table~\ref{tab:domain-adaptation} shows, the out-of-domain robustness of our instance-based models is comparable to the weight-based models.
This tendency is observed when using different source domains.

\paragraph{Sensitivity of $M$ for inference.}
In the experiments above, we used all the training sentences for support sentences at test time.
What if we reduce the number of support sentences?
Here, in the same out-of-domain settings above, we evaluate the instance-based system using the cosine similarity~\texttt{IIc} with $M$ support sentences randomly sampled at each time step.
Intuitively, if using a smaller number of randomly sampled support sentences (e.g., $M=1$), the prediction accuracies would drop.
Surprisingly, however, Table~\ref{tab:domain-adaptation-diff-M} shows that the accuracies do not drop even if reducing $M$.
This tendency is observed when using the other three systems \texttt{WId}, \texttt{WIc} and \texttt{IId}.
One possible reason for it is that the feature space is appropriately learned: i.e., because positive edges are close to each other and far from negative edges in the feature space, the accuracies do not drop even if randomly sampling a single support sentence and using the edges.

\subsection{Sanity Check for Plausible Explanations}
\label{sec:label-sim}

\begin{figure}[t]
  \begin{center}
    \includegraphics[width=7.5cm]{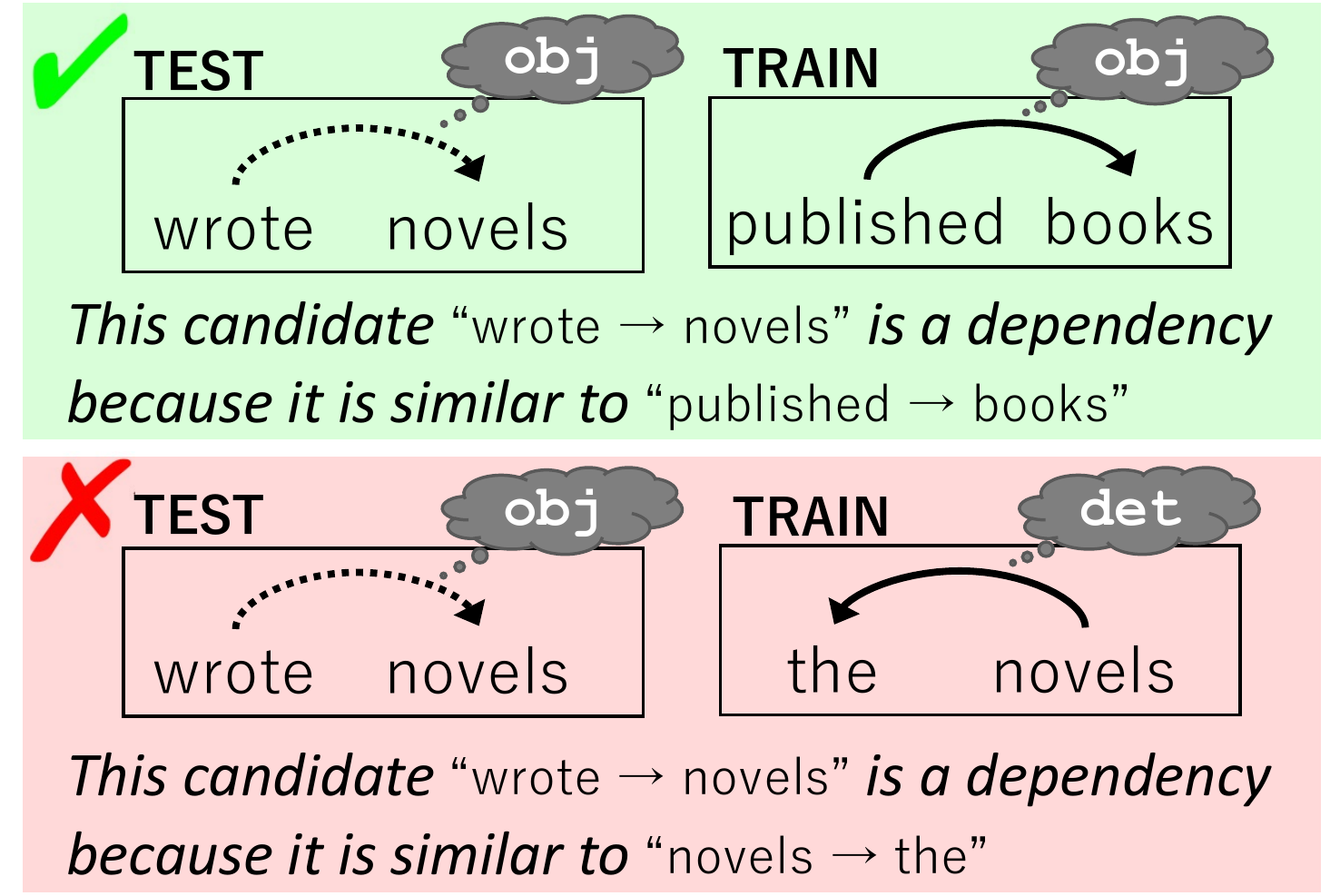}
  \end{center}
  \vspace{-0.4cm}
  \caption{Valid (\textcolor{green}{\CheckmarkBold}) and invalid (\textcolor{red}{\XSolidBrush}) examples of unlabeled edges for the \textit{identical subclass test}.}
  \label{fig:explanation}
\end{figure}

It is an open question how to evaluate the ``plausibility'' of explanations: i.e., whether or not the retrieved instances as explanations are convincing for humans.
As a reasonable compromise, \citet{hanawa2021evaluation} designed the \textbf{identical subclass test} for evaluating the plausibility.
This test is based on a minimal requirement that interpretable models should at least satisfy: \textit{training instances to be presented as explanations should belong to the same latent (sub)class as the test instance}.
Consider the examples in Figure~\ref{fig:explanation}.
The predicted unlabeled edge ``wrote $\rightarrow$ novels'' in the test sentence has the (unobserved) latent label, \texttt{obj}.
To this edge, two training instances are given as explanations: the above one seems more convincing than the below one because ``published $\rightarrow$ books'' has the same latent label, \texttt{obj}, as that of ``wrote $\rightarrow$ novels'' while ``novels $\rightarrow$ the'' has the different one, \texttt{det}.
As these show, the agreement between the latent classes are likely to correlate with plausibility.
Note that this test is not perfect for the plausibility assessment, but it works as a \textbf{sanity check} for verifying whether models make obvious violations in terms of plausibility.

This test can be used for assessing unlabeled parsing models because the (unobserved) relation labels can be regarded as the latent subclasses of positive unlabeled edges.
We follow three steps;
(i) identifying unlabeled edges in a development set;
(ii) retrieving the nearest training edge for each identified edge;
(iii) calculating LAS, i.e., if the labels of the query and retrieved edges are identical, we regard them as correct.\footnote{If the parsed edge is incorrect, we regard it as incorrect.}

\begin{table}[t]
  \centering
  {\small
  \begin{tabular}{l | rr | rr} \toprule
                 & \multicolumn{2}{c|}{Weight-Based} & \multicolumn{2}{c}{Instance-Based} \\
                 & $\mathrm{dot}$ & $\cos$ & $\mathrm{dot}$ & $\cos$ \\
                 System ID & \texttt{WId} & \texttt{WIc} & \texttt{IId} & \texttt{IIc} \\ \midrule
                 PTB-English & 1.8 & 67.5 & 7.0 & 71.6 \\
                 UD-English & 16.4 & 51.5 & 3.9 & 54.0 \\
                 \bottomrule
  \end{tabular}
  }
  \caption{Results of the identical subclass test. Each cell indicates labeled attachment scores (LAS) on each development set. All the models are trained with head selection supervision and without labeling supervision.}
  \label{tab:zero-shot}
\end{table}

Table~\ref{tab:zero-shot} shows LAS on PTB and UD-English.
The systems using instance-based inference with the cosine similarity, \texttt{WIc} and \texttt{IIc}, succeeded in retrieving the support training edges with the same label as the queries.
Surprisingly, the system \texttt{IIc} achieved over 70\% LAS on PTB without label supervision.
The results suggest that systems using instance-based inference with the cosine similarity meet the minimal requirement, and the retrieved edges are promising as plausible explanations.

\begin{table}[t]
  \centering
  {\footnotesize
  \begin{tabular}{c | c} \toprule
  Query & Support \\ \midrule
  \begin{dependency}[text only label, label style={above}, edge style={green!60!black,very thick}]
	\begin{deptext}
		If \& you \& are \& unsure \& ... \\
	\end{deptext}
	\depedge[edge unit distance=0.11cm]{4}{1}{}
  \end{dependency} &
  \begin{dependency}[text only label, label style={above}, edge style={green!60!black,very thick}]
	\begin{deptext}
		If \& you \& feel \& this \& ...  \\
	\end{deptext}
	\depedge[edge unit distance=0.15cm]{3}{1}{}
  \end{dependency} \\ \midrule
  \begin{dependency}[text only label, label style={above}, edge style={green!60!black,very thick}]
	\begin{deptext}
		... \& your \& appeal \& is \& ...  \\
	\end{deptext}
	\depedge[edge unit distance=0.3cm]{3}{2}{}
  \end{dependency} &
  \begin{dependency}[text only label, label style={above}, edge style={green!60!black,very thick}]
	\begin{deptext}
		... \& The \& food \& was \& ...  \\
	\end{deptext}
	\depedge[edge unit distance=0.3cm]{3}{2}{}
  \end{dependency}\\
  \bottomrule
  \end{tabular}
  }
  \caption{Examples of support edges retrieved by the instance-based system using the cosine similarity (\texttt{IIc}). The first query-support pair has the same (unobserved) label \texttt{mark}. The query of the second pair has \texttt{nmod:poss} although the support has \texttt{det}.}
  \label{tab:example_knn}
\end{table}

To facilitate the intuitive understanding of model behaviors, we show actual examples of the retrieved support edges in Table~\ref{tab:example_knn}.
As the first query-support pair shows, for query edges whose head or dependent is a function word (e.g., \textit{if}), the training edges with the same (unobserved) label tend to be retrieved.
On the other hand, as the second pair shows, for queries whose head is a noun (e.g., \textit{appeal}), the edges whose head is also a noun (e.g., \textit{food}) tend to be retrieved regardless of different latent labels.

\subsection{Geometric Analysis on Feature Spaces}
\label{sec:analysis}

The identical subclass test suggests a big difference between the feature spaces learned by using the dot product and the cosine similarity.
Here we look into them in more detail.

\subsubsection{Observation of Nearest Neighbors}
\label{sec:obs_nn}

\begin{table}[t]
  \centering
  {\footnotesize
  \begin{tabular}{c | c} \toprule
  Query & Support \\ \midrule
  \begin{dependency}[text only label, label style={above}, edge style={green!60!black,very thick}]
	\begin{deptext}
		Jennifer \& M. \& Anderson \& ... \\
	\end{deptext}
	\depedge[edge unit distance=0.15cm]{1}{3}{}
  \end{dependency} &
  \begin{dependency}[text only label, label style={above}, edge style={green!60!black,very thick}]
	\begin{deptext}
		\textsc{ROOT} \& Please \& find \& ...  \\
	\end{deptext}
	\depedge[edge unit distance=0.15cm]{1}{3}{}
  \end{dependency} \\ \midrule
  \begin{dependency}[text only label, label style={above}, edge style={green!60!black,very thick}]
	\begin{deptext}
		..., \& after \& all \& ...  \\
	\end{deptext}
	\depedge[edge unit distance=0.3cm]{3}{2}{}
  \end{dependency} &
  \begin{dependency}[text only label, label style={above}, edge style={green!60!black,very thick}]
	\begin{deptext}
		\textsc{ROOT} \& Please \& find \& ...  \\
	\end{deptext}
	\depedge[edge unit distance=0.15cm]{1}{3}{}
  \end{dependency} \\
  \bottomrule
  \end{tabular}
  }
  \vspace{-0.2cm}
  \caption{Examples of unlabeled support training edges retrieved by the \texttt{WId} system  (weight-based learning and instance-based inference with the dot product) for each query. Regardless of the very different queries, the same support edge was retrieved.}
  \label{tab:example:knn}
\end{table}

First, we look into training edges retrieved as nearest support ones.
Specifically, we use the edges in the UD-English development set as queries and retrieve the top $k$ similar support edges in the UD-English training set.
Table~\ref{tab:example:knn} shows the examples retrieved by the \texttt{WId} system.
Here, the same support edge, $\langle$\textit{ROOT}, \textit{find}$\rangle$, was retrieved for the different queries, $\langle$\textit{Jennifer}, \textit{Anderson}$\rangle$ and $\langle$\textit{all}, \textit{after}$\rangle$.
As this indicates, when using the dot product as the similarity function, a small number of specific edges are extremely often selected as support ones for any queries.
Such edges are called \textbf{hubs}~\cite{radovanovic2010hubs}.
This phenomenon is not desirable for users in terms of the plausible interpretation of predictions.
If a system always exhibits the same training instance(s) as rationales for predictions, users are likely to doubt the system's validity.

\subsubsection{Quantitative Measurement of Hubness}
\label{sec:hubs}

\begin{table}[t]
  \centering
  {\small
  \begin{tabular}{cc |  rc} \toprule
  		  System ID & $\mathrm{sim}$ & $N_{10}$ &  Instances \\ \midrule
		   \texttt{WId} & dot & 19,407 & $\langle$\textit{ROOT}, \textit{find}$\rangle$\\
		   \texttt{WIc} & cos & 82 & $\langle$\textit{help}, \textit{time}$\rangle$\\
		   \texttt{IId} & dot & 22,493 & $\langle$\textit{said}, \textit{airlifted}$\rangle$\\
		   \texttt{IIc} & cos & 34 & $\langle$\textit{force}, \textit{Israel}$\rangle$ \\
\bottomrule 
  \end{tabular}
  }
  \caption{Examples of the highest $N_{10}$ unlabeled support training edges in UD-English.}
  \label{tab:example:hub}
\end{table}

Second, we quantitatively measure the \textbf{hubness} of each system.
Specifically, for the hubness, we measure the $k$-occurrences of instance $x$, $N_k(x)$~\cite{radovanovic2010hubs,schnitzer2012local}. 
In the case of our dependency parsing experiments, $N_k(x)$ indicates the number of times each support training edge $x$ occurs among the $k$ nearest neighbors of all the query edges.
The support training edges with an extremely high $N_k$ value can be regarded as hubs.
In this study, we set $k=10$ and measure $N_{10}(x)$ of unlabeled support training edges.
For query edges, we use the UD-English development set that contains $25,148$ edges.
For support edges, we use the UD-English training set that contains $204,585$ edges.

\begin{figure}[t]
  \begin{center}
    \includegraphics[width=7.5cm]{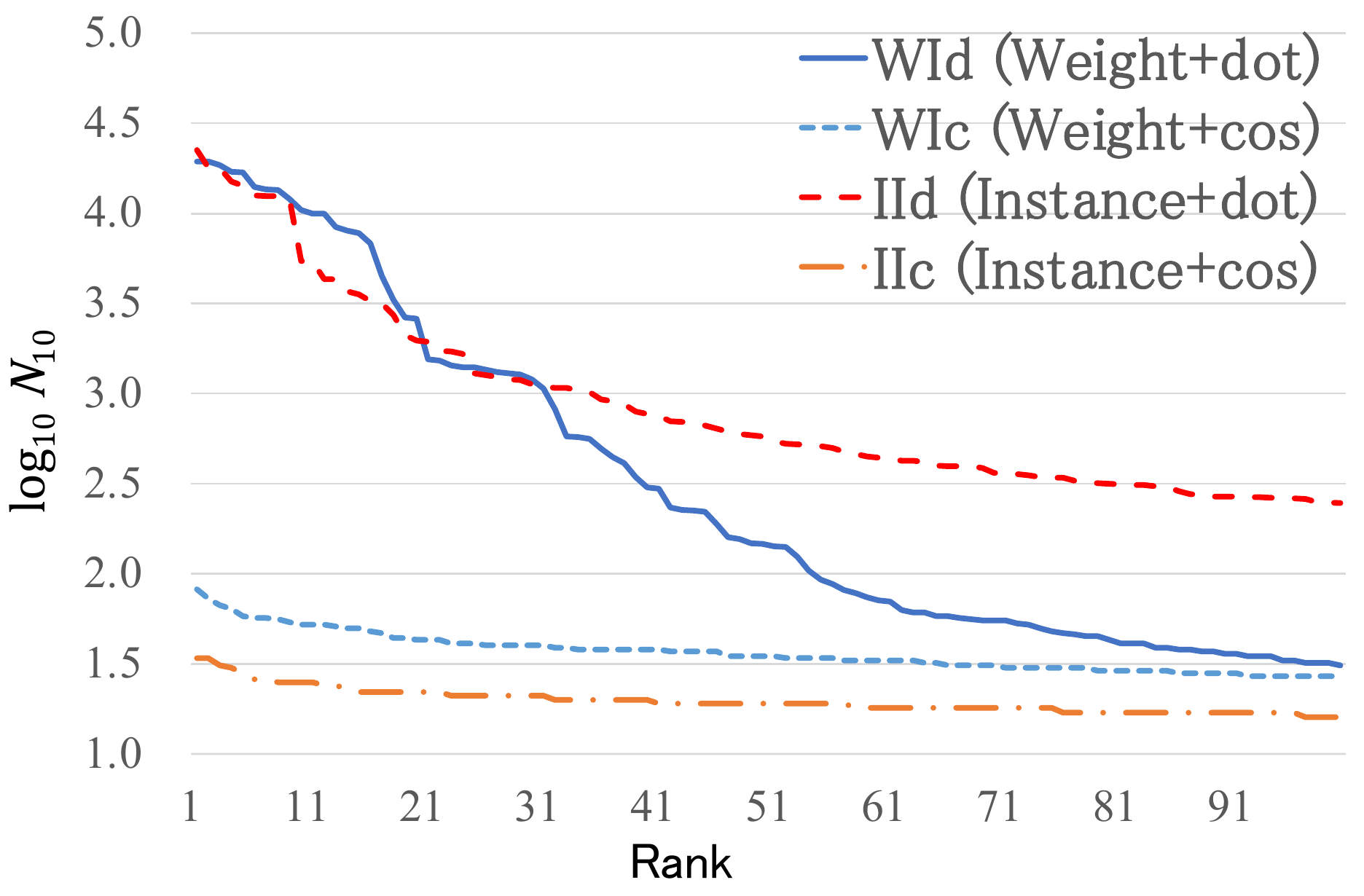}
  \end{center}
  \vspace{-0.4cm}
  \caption{Ranking of the 100 highest $N_{10}$ unlabeled support training edges in UD-English.}
  \label{fig:plot}
\end{figure}

Table~\ref{tab:example:hub} shows the highest $N_{10}$ support training edges.
In the case of the system \texttt{WId}, the unlabeled support edge $\langle$\textit{ROOT}, \textit{find}$\rangle$ appeared $19,407$ times in the $10$ nearest neighbors of the $25,148$ query edges.
A similar tendency was observed in the instance-based system using the dot product \texttt{IId}.
By contrast, in the case of the systems using the cosine similarity, \texttt{WIc} and \texttt{IIc}, it was not observed that specific support edges were retrieved so often.
In Figure~\ref{fig:plot}, we plot the top $100$ support training edges in terms of $N_{10}$ with $\log_{10}$ scale.
The $N_{10}$ distributions of the systems using the dot product, \texttt{WId} and \texttt{IId}, look very skew; that is, hubs emerge.
This indicates that when using the dot product, a small number of specific support training edges appear in the nearest neighbors so often, regardless of query edges.

To sum up, systems using instance-based inference and the dot product are often in trouble with hubs and have difficulty retrieving plausible support edges for predictions.
The occurrence of hubs are likely to be related to the norms of edge representations since L2-normalization for the edges in the cosine similarity tends to suppress hubs' occurrence.
We leave a more detailed analysis of the cause of hubs' occurrence for future work.

\section{Conclusion}
\label{sec:conc}
We have developed instance-based neural dependency parsing systems, each of which consists of our edge identification model and our label classification model (Section~\ref{sec:method}).
We have analyzed them from the perspectives of the prediction accuracy and the explanation plausibility.
The first analysis shows that our instance-based systems and achieve competitive accuracy with weight-based neural ones  (Section~\ref{sec:main-results}).
The second indicates that our instance-based systems using the cosine similarity (L2-normalization for edge representations) meet the minimal requirement of plausible explanations  (Section~\ref{sec:label-sim}).
The additional analysis reveals that when using the dot product, hubs emerge, which degrades the plausibility  (Section~\ref{sec:analysis}).
One interesting future direction is investigating the cause of hubs' occurrence in more detail.
Another direction is using the learned edge representations in downstream tasks, such as semantic textual similarity.

\section*{Acknowledgments}
The authors are grateful to the anonymous reviewers and the Action Editor who provided many insightful comments that improve the paper.
Special thanks also go to the members of Tohoku NLP Laboratory for the interesting comments and energetic discussions.
The work of H.Ouchi was supported by JSPS KAKENHI Grant Number 19K20351.
The work of J.Suzuki was supported by JST Moonshot R\&D Grant Number JPMJMS2011 (fundamental research) and JSPS KAKENHI Grant Number 19H04162.
The work of S.Yokoi was supported by JST ACT-X Grant Number JPMJAX200S, Japan.
The work of T.Kuribayashi was supported by JSPS KAKENHI Grant Number 20J22697.
The work of M.Yoshikawa was supported by JSPS KAKENHI Grant Number 20K23314.
The work of K.Inui was supported by JST CREST Grant Number JPMJCR20D2, Japan.

\bibliography{tacl2018}

\begin{thebibliography}{76}
\expandafter\ifx\csname natexlab\endcsname\relax\def\natexlab#1{#1}\fi

\bibitem[{Aha et~al.(1991)Aha, Kibler, and Albert}]{aha1991instance}
David~W Aha, Dennis Kibler, and Marc~K Albert. 1991.
\newblock Instance-based learning algorithms.
\newblock \emph{Machine learning}, 6(1):37--66.

\bibitem[{Akbik and Li(2016)}]{akbik-li-2016-k}
Alan Akbik and Yunyao Li. 2016.
\newblock \href {https://www.aclweb.org/anthology/C16-1058} {K-{SRL}:
  Instance-based learning for semantic role labeling}.
\newblock In \emph{Proceedings of {COLING}}, pages 599--608.

\bibitem[{Baehrens et~al.(2010)Baehrens, Schroeter, Harmeling, Kawanabe,
  Hansen, and M{\~A}{\v{z}}ller}]{baehrens2010explain}
David Baehrens, Timon Schroeter, Stefan Harmeling, Motoaki Kawanabe, Katja
  Hansen, and Klaus-Robert M{\~A}{\v{z}}ller. 2010.
\newblock How to explain individual classification decisions.
\newblock \emph{Journal of Machine Learning Research}, 11(Jun):1803--1831.

\bibitem[{Bellet et~al.(2013)Bellet, Habrard, and Sebban}]{bellet2013survey}
Aur{\'e}lien Bellet, Amaury Habrard, and Marc Sebban. 2013.
\newblock A survey on metric learning for feature vectors and structured data.
\newblock \textit{arXiv preprint arXiv:1306.6709}.

\bibitem[{Bod(2009)}]{bod2009exemplar}
Rens Bod. 2009.
\newblock From exemplar to grammar: A probabilistic analogy-based model of
  language learning.
\newblock \emph{Cognitive Science}, 33(5):752--793.

\bibitem[{Bordes et~al.(2013)Bordes, Usunier, Garcia-Duran, Weston, and
  Yakhnenko}]{bordes2013translating}
Antoine Bordes, Nicolas Usunier, Alberto Garcia-Duran, Jason Weston, and Oksana
  Yakhnenko. 2013.
\newblock Translating embeddings for modeling multi-relational data.
\newblock \emph{Proceedings of NIPS}, 26:2787--2795.

\bibitem[{Caruana et~al.(1999)Caruana, Kangarloo, Dionisio, Sinha, and
  Johnson}]{caruana1999case}
Rich Caruana, Hooshang Kangarloo, John~David Dionisio, Usha Sinha, and David
  Johnson. 1999.
\newblock Case-based explanation of non-case-based learning methods.
\newblock In \emph{Proceedings of the AMIA Symposium}, page 212.

\bibitem[{Clark et~al.(2018)Clark, Luong, Manning, and
  Le}]{clark-etal-2018-semi}
Kevin Clark, Minh-Thang Luong, Christopher~D. Manning, and Quoc Le. 2018.
\newblock \href {https://doi.org/10.18653/v1/D18-1217} {Semi-supervised
  sequence modeling with cross-view training}.
\newblock In \emph{Proceedings of EMNLP}, pages 1914--1925.

\bibitem[{Daelemans and Van~den Bosch(2005)}]{daelemans2005memory}
Walter Daelemans and Antal Van~den Bosch. 2005.
\newblock \emph{Memory-based language processing}.
\newblock Cambridge University Press.

\bibitem[{Daelemans et~al.(1999)Daelemans, Buchholz, and
  Veenstra}]{daelemans1999memory}
Walter Daelemans, Sabine Buchholz, and Jorn Veenstra. 1999.
\newblock Memory-based shallow parsing.
\newblock In \emph{EACL 1999: CoNLL-99 Computational Natural Language
  Learning}.

\bibitem[{Daelemans et~al.(1996)Daelemans, Zavrel, Berck, and
  Gillis}]{daelemans-etal-1996-mbt}
Walter Daelemans, Jakub Zavrel, Peter Berck, and Steven Gillis. 1996.
\newblock \href {https://www.aclweb.org/anthology/W96-0102} {{MBT}: A
  memory-based part of speech tagger-generator}.
\newblock In \emph{Proceedings of Fourth Workshop on Very Large Corpora}.

\bibitem[{De~Meulder and Daelemans(2003)}]{de-meulder-daelemans-2003-memory}
Fien De~Meulder and Walter Daelemans. 2003.
\newblock \href {https://www.aclweb.org/anthology/W03-0435} {Memory-based named
  entity recognition using unannotated data}.
\newblock In \emph{Proceedings of HLT-NAACL}, pages 208--211.

\bibitem[{Deng et~al.(2019)Deng, Guo, Xue, and Zafeiriou}]{deng2019arcface}
Jiankang Deng, Jia Guo, Niannan Xue, and Stefanos Zafeiriou. 2019.
\newblock Arcface: Additive angular margin loss for deep face recognition.
\newblock In \emph{Proceedings of CVPR}, pages 4690--4699.

\bibitem[{Devlin et~al.(2019)Devlin, Chang, Lee, and
  Toutanova}]{devlin-etal-2019-bert}
Jacob Devlin, Ming-Wei Chang, Kenton Lee, and Kristina Toutanova. 2019.
\newblock \href {https://doi.org/10.18653/v1/N19-1423} {{BERT}: Pre-training of
  deep bidirectional transformers for language understanding}.
\newblock In \emph{Proceedings of NAACL-HLT}, pages 4171--4186.

\bibitem[{Dozat and Manning(2017)}]{dozat2017deep}
Timothy Dozat and Christopher~D Manning. 2017.
\newblock Deep biaffine attention for neural dependency parsing.
\newblock In \emph{Proceedings of ICLR}.

\bibitem[{Friedman et~al.(1994)}]{friedman1994flexible}
Jerome~H Friedman et~al. 1994.
\newblock Flexible metric nearest neighbor classification.
\newblock Technical report.

\bibitem[{Goldberger et~al.(2005)Goldberger, Hinton, Roweis, and
  Salakhutdinov}]{goldberger2005neighbourhood}
Jacob Goldberger, Geoffrey~E Hinton, Sam~T Roweis, and Ruslan~R Salakhutdinov.
  2005.
\newblock \href {https://www.cs.toronto.edu/~hinton/absps/nca.pdf}
  {Neighbourhood components analysis}.
\newblock In \emph{Proceedings of NIPS}, pages 513--520.

\bibitem[{Grave et~al.(2018)Grave, Bojanowski, Gupta, Joulin, and
  Mikolov}]{grave2018learning}
Edouard Grave, Piotr Bojanowski, Prakhar Gupta, Armand Joulin, and Tomas
  Mikolov. 2018.
\newblock Learning word vectors for 157 languages.
\newblock In \emph{Proceedings of LREC}.

\bibitem[{Graves et~al.(2013)Graves, Jaitly, and Mohamed}]{graves:13}
Alan Graves, Navdeep Jaitly, and Abdel-rahman Mohamed. 2013.
\newblock Hybrid speech recognition with deep bidirectional {LSTM}.
\newblock In \emph{Proceedings of Automatic Speech Recognition and
  Understanding (ASRU), 2013 IEEE Workshop}.

\bibitem[{Guu et~al.(2020)Guu, Lee, Tung, Pasupat, and Chang}]{guu2020realm}
Kelvin Guu, Kenton Lee, Zora Tung, Panupong Pasupat, and Ming-Wei Chang. 2020.
\newblock Realm: Retrieval-augmented language model pre-training.
\newblock \emph{arXiv preprint arXiv:2002.08909}.

\bibitem[{Hadsell et~al.(2006)Hadsell, Chopra, and
  LeCun}]{hadsell2006dimensionality}
Raia Hadsell, Sumit Chopra, and Yann LeCun. 2006.
\newblock Dimensionality reduction by learning an invariant mapping.
\newblock In \emph{Proceedings of CVPR}, volume~2, pages 1735--1742. IEEE.

\bibitem[{Hanawa et~al.(2021)Hanawa, Yokoi, Hara, and
  Inui}]{hanawa2021evaluation}
Kazuaki Hanawa, Sho Yokoi, Satoshi Hara, and Kentaro Inui. 2021.
\newblock \href {https://openreview.net/forum?id=9uvhpyQwzM_} {Evaluation of
  similarity-based explanations}.
\newblock In \emph{Proceedings of ICLR}.

\bibitem[{Hashimoto et~al.(2017)Hashimoto, Xiong, Tsuruoka, and
  Socher}]{hashimoto-etal-2017-joint}
Kazuma Hashimoto, Caiming Xiong, Yoshimasa Tsuruoka, and Richard Socher. 2017.
\newblock \href {https://doi.org/10.18653/v1/D17-1206} {A joint many-task
  model: Growing a neural network for multiple {NLP} tasks}.
\newblock In \emph{Proceedings of EMNLP}, pages 1923--1933.

\bibitem[{Haspelmath et~al.(2005)Haspelmath, Dryer, Gil, and
  Comrie}]{haspelmath2005world}
Martin Haspelmath, Matthew~S Dryer, David Gil, and Bernard Comrie. 2005.
\newblock The world atlas of language structures.

\bibitem[{Hastie and Tibshirani(1996)}]{hastie1996discriminant}
Trevor Hastie and Robert Tibshirani. 1996.
\newblock Discriminant adaptive nearest neighbor classification and regression.
\newblock In \emph{Proceedings of NIPS}, pages 409--415.

\bibitem[{Hendrickx and van~den
  Bosch(2003)}]{hendrickx-van-den-bosch-2003-memory}
Iris Hendrickx and Antal van~den Bosch. 2003.
\newblock \href {https://www.aclweb.org/anthology/W03-0427} {Memory-based
  one-step named-entity recognition: Effects of seed list features, classifier
  stacking, and unannotated data}.
\newblock In \emph{Proceedings of CoNLL}, pages 176--179.

\bibitem[{Hoffer and Ailon(2015)}]{hoffer2015deep}
Elad Hoffer and Nir Ailon. 2015.
\newblock Deep metric learning using triplet network.
\newblock In \emph{International Workshop on Similarity-Based Pattern
  Recognition}, pages 84--92. Springer.

\bibitem[{Ji et~al.(2019)Ji, Wu, and Lan}]{ji-etal-2019-graph}
Tao Ji, Yuanbin Wu, and Man Lan. 2019.
\newblock \href {https://doi.org/10.18653/v1/P19-1237} {Graph-based dependency
  parsing with graph neural networks}.
\newblock In \emph{Proceedings of ACL}, pages 2475--2485.

\bibitem[{Khandelwal et~al.(2020)Khandelwal, Fan, Jurafsky, Zettlemoyer, and
  Lewis}]{khandelwal2020nearest}
Urvashi Khandelwal, Angela Fan, Dan Jurafsky, Luke Zettlemoyer, and Mike Lewis.
  2020.
\newblock Nearest neighbor machine translation.
\newblock \textit{arXiv preprint arXiv:2010.00710}.

\bibitem[{Khandelwal et~al.(2019)Khandelwal, Levy, Jurafsky, Zettlemoyer, and
  Lewis}]{khandelwal2019generalization}
Urvashi Khandelwal, Omer Levy, Dan Jurafsky, Luke Zettlemoyer, and Mike Lewis.
  2019.
\newblock Generalization through memorization: Nearest neighbor language
  models.
\newblock In \emph{Proceedings of ICLR}.

\bibitem[{Kingma and Ba(2014)}]{kingma:14}
D.P. Kingma and J.~Ba. 2014.
\newblock Adam: A method for stochastic optimization.
\newblock \textit{arXiv preprint arXiv:1412.6980}.

\bibitem[{Kolodneer(1991)}]{kolodneer1991improving}
Janet~L Kolodneer. 1991.
\newblock Improving human decision making through case-based decision aiding.
\newblock \emph{AI magazine}, 12(2):52--52.

\bibitem[{K{\"u}bler(2004)}]{kubler2004memory}
Sandra K{\"u}bler. 2004.
\newblock \emph{Memory-based parsing}, volume~7.

\bibitem[{K{\"u}bler et~al.(2009)K{\"u}bler, McDonald, and
  Nivre}]{kubler2009dependency}
Sandra K{\"u}bler, Ryan McDonald, and Joakim Nivre. 2009.
\newblock Dependency parsing.
\newblock \emph{Synthesis lectures on human language technologies},
  1(1):1--127.

\bibitem[{Kulmizev et~al.(2019)Kulmizev, de~Lhoneux, Gontrum, Fano, and
  Nivre}]{kulmizev2019deep}
Artur Kulmizev, Miryam de~Lhoneux, Johannes Gontrum, Elena Fano, and Joakim
  Nivre. 2019.
\newblock Deep contextualized word embeddings in transition-based and
  graph-based dependency parsing-a tale of two parsers revisited.
\newblock In \emph{Proceedings of EMNLP-IJCNLP}, pages 2755--2768.

\bibitem[{Lebowitz(1983)}]{lebowitz1983memory}
Michael Lebowitz. 1983.
\newblock Memory-based parsing.
\newblock \emph{Artificial Intelligence}, 21(4):363--404.

\bibitem[{Lei et~al.(2016)Lei, Barzilay, and
  Jaakkola}]{lei-etal-2016-rationalizing}
Tao Lei, Regina Barzilay, and Tommi Jaakkola. 2016.
\newblock \href {https://doi.org/10.18653/v1/D16-1011} {Rationalizing neural
  predictions}.
\newblock In \emph{Proceedings of EMNLP}, pages 107--117.

\bibitem[{Lewis et~al.(2020)Lewis, Perez, Piktus, Petroni, Karpukhin, Goyal,
  K{\"u}ttler, Lewis, Yih, Rockt{\"a}schel et~al.}]{lewis2020retrieval}
Patrick Lewis, Ethan Perez, Aleksandara Piktus, Fabio Petroni, Vladimir
  Karpukhin, Naman Goyal, Heinrich K{\"u}ttler, Mike Lewis, Wen-tau Yih, Tim
  Rockt{\"a}schel, et~al. 2020.
\newblock Retrieval-augmented generation for knowledge-intensive nlp tasks.
\newblock \emph{arXiv preprint arXiv:2005.11401}.

\bibitem[{de~Lhoneux et~al.(2017)de~Lhoneux, Stymne, and Nivre}]{de2017old}
Miryam de~Lhoneux, Sara Stymne, and Joakim Nivre. 2017.
\newblock Old school vs. new school: Comparing transition-based parsers with
  and without neural network enhancement.
\newblock In \emph{The 15th Treebanks and Linguistic Theories Workshop (TLT)},
  pages 99--110.

\bibitem[{Liu et~al.(2017)Liu, Wen, Yu, Li, Raj, and Song}]{liu2017sphereface}
Weiyang Liu, Yandong Wen, Zhiding Yu, Ming Li, Bhiksha Raj, and Le~Song. 2017.
\newblock Sphereface: Deep hypersphere embedding for face recognition.
\newblock In \emph{Proceedings of CVPR}, pages 212--220.

\bibitem[{Marcus et~al.(1993)Marcus, Santorini, and
  Marcinkiewicz}]{marcus-etal-1993-building}
Mitchell~P. Marcus, Beatrice Santorini, and Mary~Ann Marcinkiewicz. 1993.
\newblock \href {https://www.aclweb.org/anthology/J93-2004} {Building a large
  annotated corpus of {E}nglish: The {P}enn {T}reebank}.
\newblock \emph{Computational Linguistics}, 19(2):313--330.

\bibitem[{McDonald et~al.(2006)McDonald, Lerman, and
  Pereira}]{mcdonald-etal-2006-multilingual}
Ryan McDonald, Kevin Lerman, and Fernando Pereira. 2006.
\newblock \href {https://www.aclweb.org/anthology/W06-2932} {Multilingual
  dependency analysis with a two-stage discriminative parser}.
\newblock In \emph{Proceedings of {C}o{NLL}-X}, pages 216--220.

\bibitem[{McDonald et~al.(2013)McDonald, Nivre, Quirmbach-Brundage, Goldberg,
  Das, Ganchev, Hall, Petrov, Zhang, T{\"a}ckstr{\"o}m, Bedini,
  Bertomeu~Castell{\'o}, and Lee}]{mcdonald-etal-2013-universal}
Ryan McDonald, Joakim Nivre, Yvonne Quirmbach-Brundage, Yoav Goldberg, Dipanjan
  Das, Kuzman Ganchev, Keith Hall, Slav Petrov, Hao Zhang, Oscar
  T{\"a}ckstr{\"o}m, Claudia Bedini, N{\'u}ria Bertomeu~Castell{\'o}, and
  Jungmee Lee. 2013.
\newblock \href {https://www.aclweb.org/anthology/P13-2017} {{U}niversal
  {D}ependency annotation for multilingual parsing}.
\newblock In \emph{Proceedings of ACL}, pages 92--97.

\bibitem[{McDonald et~al.(2005)McDonald, Pereira, Ribarov, and
  Haji{\v{c}}}]{mcdonald-etal-2005-non}
Ryan McDonald, Fernando Pereira, Kiril Ribarov, and Jan Haji{\v{c}}. 2005.
\newblock \href {https://www.aclweb.org/anthology/H05-1066} {Non-projective
  dependency parsing using spanning tree algorithms}.
\newblock In \emph{Proceedings of HLT-EMNLP}, pages 523--530.

\bibitem[{Nagao(1984)}]{nagao1984framework}
Makoto Nagao. 1984.
\newblock \href {http://www.mt-archive.info/Nagao-1984.pdf} {\emph{A framework
  of a mechanical translation between Japanese and English by analogy
  principle}}.

\bibitem[{Nickel et~al.(2016)Nickel, Rosasco, and
  Poggio}]{nickel2016holographic}
Maximilian Nickel, Lorenzo Rosasco, and Tomaso Poggio. 2016.
\newblock Holographic embeddings of knowledge graphs.
\newblock In \emph{Proceedings of AAAI}, volume~30.

\bibitem[{Nivre(2003)}]{nivre2003efficient}
Joakim Nivre. 2003.
\newblock An efficient algorithm for projective dependency parsing.
\newblock In \emph{Proceedings of the eighth international conference on
  parsing technologies}, pages 149--160.

\bibitem[{Nivre et~al.(2004)Nivre, Hall, and Nilsson}]{nivre-etal-2004-memory}
Joakim Nivre, Johan Hall, and Jens Nilsson. 2004.
\newblock \href {https://www.aclweb.org/anthology/W04-2407} {Memory-based
  dependency parsing}.
\newblock In \emph{Proceedings of {C}o{NLL}}, pages 49--56.

\bibitem[{Oord et~al.(2018)Oord, Li, and Vinyals}]{oord2018representation}
Aaron van~den Oord, Yazhe Li, and Oriol Vinyals. 2018.
\newblock Representation learning with contrastive predictive coding.
\newblock \emph{arXiv preprint arXiv:1807.03748}.

\bibitem[{Ouchi et~al.(2020)Ouchi, Suzuki, Kobayashi, Yokoi, Kuribayashi,
  Konno, and Inui}]{ouchi-etal-2020-instance}
Hiroki Ouchi, Jun Suzuki, Sosuke Kobayashi, Sho Yokoi, Tatsuki Kuribayashi,
  Ryuto Konno, and Kentaro Inui. 2020.
\newblock \href {https://doi.org/10.18653/v1/2020.acl-main.575} {Instance-based
  learning of span representations: A case study through named entity
  recognition}.
\newblock In \emph{Proceedings of ACL}, pages 6452--6459.

\bibitem[{Papernot and McDaniel(2018)}]{papernot2018deep}
Nicolas Papernot and Patrick McDaniel. 2018.
\newblock Deep k-nearest neighbors: Towards confident, interpretable and robust
  deep learning.
\newblock \emph{arXiv preprint arXiv:1803.04765}.

\bibitem[{Pascanu et~al.(2013)Pascanu, Mikolov, and
  Bengio}]{pascanu2013difficulty}
Razvan Pascanu, Tomas Mikolov, and Yoshua Bengio. 2013.
\newblock On the difficulty of training recurrent neural networks.
\newblock In \emph{Proceedings of ICML}, pages 1310--1318.

\bibitem[{Pennington et~al.(2014)Pennington, Socher, and
  Manning}]{pennington-etal-2014-glove}
Jeffrey Pennington, Richard Socher, and Christopher Manning. 2014.
\newblock \href {https://doi.org/10.3115/v1/D14-1162} {{G}love: Global vectors
  for word representation}.
\newblock In \emph{Proceedings of EMNLP}, pages 1532--1543.

\bibitem[{Plumb et~al.(2018)Plumb, Molitor, and Talwalkar}]{plumb2018model}
Gregory Plumb, Denali Molitor, and Ameet~S Talwalkar. 2018.
\newblock Model agnostic supervised local explanations.
\newblock In \emph{Proceedings of NIPS}, pages 2515--2524.

\bibitem[{Radovanovic et~al.(2010)Radovanovic, Nanopoulos, and
  Ivanovic}]{radovanovic2010hubs}
Milos Radovanovic, Alexandros Nanopoulos, and Mirjana Ivanovic. 2010.
\newblock Hubs in space: Popular nearest neighbors in high-dimensional data.
\newblock \emph{Journal of Machine Learning Research}, 11(sept):2487--2531.

\bibitem[{Ranjan et~al.(2017)Ranjan, Castillo, and Chellappa}]{ranjan2017l2}
Rajeev Ranjan, Carlos~D Castillo, and Rama Chellappa. 2017.
\newblock L2-constrained softmax loss for discriminative face verification.
\newblock \emph{arXiv preprint arXiv:1703.09507}.

\bibitem[{Ribeiro et~al.(2016)Ribeiro, Singh, and Guestrin}]{ribeiro2016should}
Marco~Tulio Ribeiro, Sameer Singh, and Carlos Guestrin. 2016.
\newblock Why should i trust you?: Explaining the predictions of any
  classifier.
\newblock In \emph{Proceedings of KDD}, pages 1135--1144.

\bibitem[{Sang(2002)}]{sang2002memory}
Erik F Tjong~Kim Sang. 2002.
\newblock Memory-based shallow parsing.
\newblock \emph{Journal of Machine Learning Research}, 2:559--594.

\bibitem[{Scha et~al.(1999)Scha, Bod, and Sima'An}]{scha1999memory}
Remko Scha, Rens Bod, and Khalil Sima'An. 1999.
\newblock A memory-based model of syntactic analysis: data-oriented parsing.
\newblock \emph{Journal of Experimental \& Theoretical Artificial
  Intelligence}, 11(3):409--440.

\bibitem[{Schnitzer et~al.(2012)Schnitzer, Flexer, Schedl, and
  Widmer}]{schnitzer2012local}
Dominik Schnitzer, Arthur Flexer, Markus Schedl, and Gerhard Widmer. 2012.
\newblock Local and global scaling reduce hubs in space.
\newblock \emph{Journal of Machine Learning Research}, 13(10).

\bibitem[{Short and Fukunaga(1981)}]{short1981optimal}
R~Short and Keinosuke Fukunaga. 1981.
\newblock The optimal distance measure for nearest neighbor classification.
\newblock \emph{IEEE transactions on Information Theory}, 27(5):622--627.

\bibitem[{Smith et~al.(2018)Smith, de~Lhoneux, Stymne, and
  Nivre}]{smith2018investigation}
Aaron Smith, Miryam de~Lhoneux, Sara Stymne, and Joakim Nivre. 2018.
\newblock An investigation of the interactions between pre-trained word
  embeddings, character models and pos tags in dependency parsing.
\newblock In \emph{Proceedings of EMNLP}, pages 2711--2720.

\bibitem[{Sohn(2016)}]{sohn2016improved}
Kihyuk Sohn. 2016.
\newblock Improved deep metric learning with multi-class n-pair loss objective.
\newblock In \emph{Proceedings of NIPS}, pages 1857--1865.

\bibitem[{Sumita and Iida(1991)}]{sumita-iida-1991-experiments}
Eiichiro Sumita and Hitoshi Iida. 1991.
\newblock \href {https://doi.org/10.3115/981344.981368} {Experiments and
  prospects of example-based machine translation}.
\newblock In \emph{Proceedings of ACL}, pages 185--192.

\bibitem[{Sun et~al.(2020)Sun, Cheng, Zhang, Zhang, Zheng, Wang, and
  Wei}]{sun2020circle}
Yifan Sun, Changmao Cheng, Yuhan Zhang, Chi Zhang, Liang Zheng, Zhongdao Wang,
  and Yichen Wei. 2020.
\newblock Circle loss: A unified perspective of pair similarity optimization.
\newblock In \emph{Proceedings of CVPR}, pages 6398--6407.

\bibitem[{Tjong Kim~Sang(2002)}]{tjong-kim-sang-2002-memory}
Erik~F. Tjong Kim~Sang. 2002.
\newblock \href {https://www.aclweb.org/anthology/W02-2025} {Memory-based named
  entity recognition}.
\newblock In \emph{Proceedings of CoNLL}.

\bibitem[{Veenstra et~al.(2000)Veenstra, Van~den Bosch, Buchholz, Daelemans
  et~al.}]{veenstra2000memory}
Jorn Veenstra, Antal Van~den Bosch, Sabine Buchholz, Walter Daelemans, et~al.
  2000.
\newblock Memory-based word sense disambiguation.
\newblock \emph{Computers and the Humanities}, 34(1):171--177.

\bibitem[{Wang et~al.(2018)Wang, Wang, Zhou, Ji, Gong, Zhou, Li, and
  Liu}]{wang2018cosface}
Hao Wang, Yitong Wang, Zheng Zhou, Xing Ji, Dihong Gong, Jingchao Zhou, Zhifeng
  Li, and Wei Liu. 2018.
\newblock Cosface: Large margin cosine loss for deep face recognition.
\newblock In \emph{Proceedings of CVPR}, pages 5265--5274.

\bibitem[{Wang et~al.(2014)Wang, Song, Leung, Rosenberg, Wang, Philbin, Chen,
  and Wu}]{wang2014learning}
Jiang Wang, Yang Song, Thomas Leung, Chuck Rosenberg, Jingbin Wang, James
  Philbin, Bo~Chen, and Ying Wu. 2014.
\newblock Learning fine-grained image similarity with deep ranking.
\newblock In \emph{Proceedings of CVPR}, pages 1386--1393.

\bibitem[{Wang et~al.(2019)Wang, Han, Huang, Dong, and Scott}]{wang2019multi}
Xun Wang, Xintong Han, Weilin Huang, Dengke Dong, and Matthew~R Scott. 2019.
\newblock Multi-similarity loss with general pair weighting for deep metric
  learning.
\newblock In \emph{Proceedings of CVPR}, pages 5022--5030.

\bibitem[{Wiseman and Stratos(2019)}]{wiseman-stratos-2019-label}
Sam Wiseman and Karl Stratos. 2019.
\newblock \href {https://www.aclweb.org/anthology/P19-1533} {Label-agnostic
  sequence labeling by copying nearest neighbors}.
\newblock In \emph{Proceedings of ACL}, pages 5363--5369.

\bibitem[{Xing et~al.(2002)Xing, Jordan, Russell, and Ng}]{xing2002distance}
Eric Xing, Michael Jordan, Stuart~J Russell, and Andrew Ng. 2002.
\newblock Distance metric learning with application to clustering with
  side-information.
\newblock In \emph{Proceedings of NIPS}, volume~15, pages 521--528.

\bibitem[{Yamada and Matsumoto(2003)}]{yamada2003statistical}
Hiroyasu Yamada and Yuji Matsumoto. 2003.
\newblock Statistical dependency analysis with support vector machines.
\newblock In \emph{Proceedings of the eighth international conference on
  parsing technologies}, pages 195--206.

\bibitem[{Yang et~al.(2015)Yang, Yih, He, Gao, and Deng}]{yang2014embedding}
Bishan Yang, Wen-tau Yih, Xiaodong He, Jianfeng Gao, and Li~Deng. 2015.
\newblock Embedding entities and relations for learning and inference in
  knowledge bases.

\bibitem[{Zhang et~al.(2017)Zhang, Cheng, and
  Lapata}]{zhang-etal-2017-dependency-parsing}
Xingxing Zhang, Jianpeng Cheng, and Mirella Lapata. 2017.
\newblock \href {https://www.aclweb.org/anthology/E17-1063} {Dependency parsing
  as head selection}.
\newblock In \emph{Proceedings of EACL}, pages 665--676.

\bibitem[{Zhang et~al.(2020)Zhang, Li, and Zhang}]{zhang-etal-2020-efficient}
Yu~Zhang, Zhenghua Li, and Min Zhang. 2020.
\newblock \href {https://doi.org/10.18653/v1/2020.acl-main.302} {Efficient
  second-order {T}ree{CRF} for neural dependency parsing}.
\newblock In \emph{Proceedings of ACL}, pages 3295--3305.

\end{thebibliography}
\bibliographystyle{acl_natbib}

\end{document}